\documentclass{article}
\usepackage{spconf,amsmath,graphicx,hyperref}
\usepackage{booktabs}  
\usepackage{graphicx}  
\usepackage{siunitx}
\usepackage{bm}
\usepackage{afterpage}
\usepackage{float}
\usepackage{amsmath,amssymb}

\title{DiffusionX: Efficient Edge-Cloud Collaborative Image Generation with Multi-Round Prompt Evolution}
\name{Yi Wei, Shunpu Tang, Liang Zhao, Qianqian Yang}
\address{College of Information Science and Electronic Engineering, Zhejiang University, Hangzhou, China}
\begin{document}
%
\maketitle
\begin{abstract}
Recent advances in diffusion models have driven remarkable progress in image generation. However, the generation process remains computationally intensive, and users often need to iteratively refine prompts to achieve the desired results, further increasing latency and placing a heavy burden on cloud resources. To address this challenge, we propose \textit{DiffusionX}, a cloud–edge collaborative framework for efficient multi-round, prompt-based generation. In this system, a lightweight on-device diffusion model interacts with users by rapidly producing preview images, while a high-capacity cloud model performs final refinements after the prompt is finalized. We further introduce a noise level predictor that dynamically balances the computation load, optimizing the trade-off between latency and cloud workload. Experiments show that \textit{DiffusionX} reduces average generation time by 15.8\% compared with Stable Diffusion v1.5, while maintaining comparable image quality. Moreover, it is only 0.9\% slower than Tiny-SD with significantly improved image quality, yet delivers significantly better image quality, thereby demonstrating efficiency and scalability with minimal overhead.

\end{abstract}
\section{Introduction}
\label{sec:intro}

Recent advances in generative diffusion models (GDMs) have significantly improved the quality and diversity of text-to-image generation \cite{T2I_survey,DreamBooth,Rectified_Flow_Transformers}. However, these models rely on iterative denoising across hundreds of steps, and their large parameter sizes also increase computational demands. For example, the Stable Diffusion XL (SDXL) base model \cite{SDXL} contains 3.5 billion parameters and requires roughly 10 seconds to generate a 1024×1024 image on a modern GPU,  which limits its practicality for real-world deployment.

To address this inefficiency, prior works have explored accelerating generation by reducing the complexity of inference. On one hand, some studies focus on model compression techniques such as pruning and quantization \cite{kwonHierarchicalPrunePositionAwareCompression2025}, which eliminate redundant parameters and reduce per-step computation while preserving generation quality. On the other hand, researchers have investigated methods to decrease the number of required denoising steps. For example, Xia et al. \cite{c33} proposed a timestep tuner that adaptively adjusts integration directions, mitigating truncation errors and improving quality with fewer steps. The authors in \cite{c34} introduced an optimal linear subspace search (OLSS) scheduler that approximates the full process in fewer steps, enabling near real-time synthesis on powerful hardware. Moreover, the authors in \cite{c35} proposed \textit{DeepCache}, which caches intermediate features across denoising stages, effectively skipping redundant steps and achieving more than 2× speedup without retraining.

\begin{figure}[t]
  \centering
  \includegraphics[width=\columnwidth]{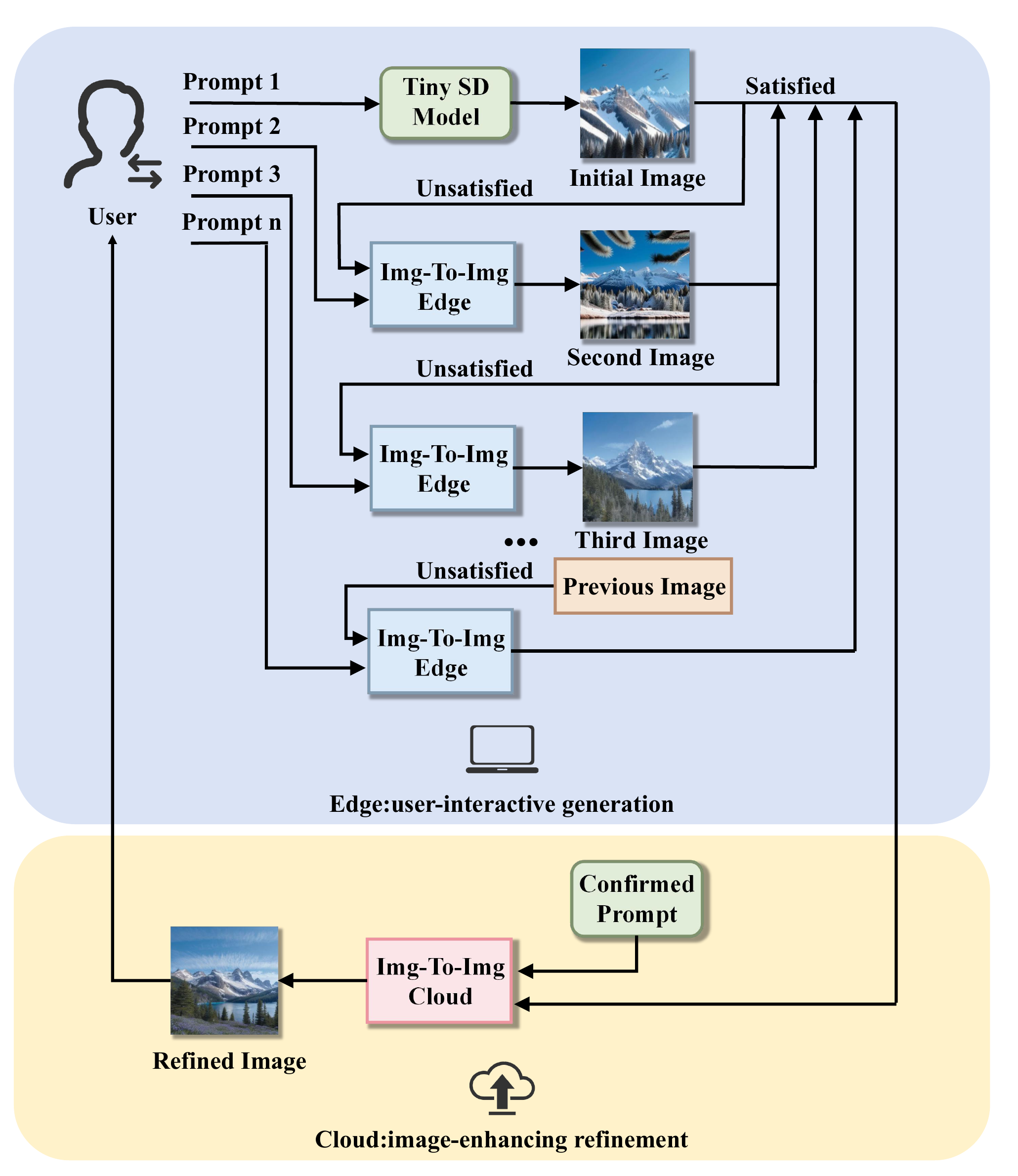}
  \caption{Illustration of the proposed \textit{DiffusionX} framework, where the edge model provides fast previews for user interaction and the cloud model refines them into high-fidelity results. \vspace{-1.2em}}
  \label{fig:system_overview}
\end{figure}

While these approaches can effectively reduce image generation latency, they still face several limitations. In practice, users often cannot obtain the desired result in a single attempt and must refine or supplement prompts based on previously generated images. However, existing techniques treat each round of generation as an independent process, leading to unnecessary system overhead. Furthermore, most approaches focus solely on the cloud-computing paradigm and overlook the potential of edge devices. Although lightweight models deployed on the edge have limited capacity, they enable opportunities for collaborative computing, where edge models generate coarse outputs and cloud models subsequently refine or validate them \cite{Hybrid_LLM,c7}.

Motivated by this, we explore efficient edge-cloud collaborative image generation by leveraging the integration of lightweight GDMs, such as Tiny SD \cite{c32}, on the edge with a large GDM in the cloud. In this setting, the edge side supports user interaction and provides coarse previews, while the cloud refines them into high-quality results. The main contributions of this paper are as follows: 1) we propose \textit{DiffusionX}, a hybrid framework that integrates a lightweight edge GDM with a large cloud GDM to better support multi-round user interaction. 2) We introduce strength predictors to reduce redundant noise estimation, thereby lowering system overhead and accelerating iterative refinement. 3) We conduct extensive experiments showing that \textit{DiffusionX} reduces generation time by 23.2\% compared with a cloud-only large GDM, while being only 2.1\% slower than the lightweight baseline, while maintaining comparable image quality to the large GDM.


\section{Proposed System}
\label{sec:method}
As shown in \autoref{fig:system_overview}, we propose \textit{DiffusionX}, a collaborative edge–cloud framework where the edge produces fast previews for user interaction and the cloud refines them into high-fidelity results. The system consists of two key modules: (1) a lightweight GDM with a semantic-aware strength predictor for fast previews; (2) a cloud-based high-fidelity predictor with skip-step denoising to refine results efficiently. 


\begin{figure}[t]
  \centering
  \includegraphics[width=\columnwidth]{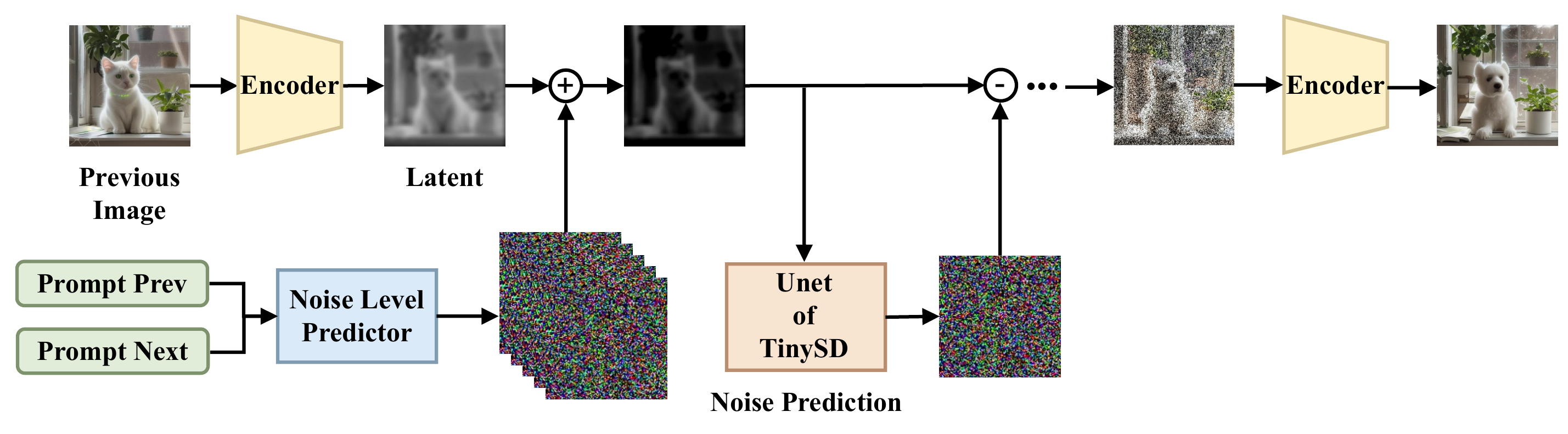}
  \caption{The structure of img2img generation on the edge with a noise level predictor.}
  \label{fig:img2img_edge}
\end{figure}

\vspace{-0.5em}
\subsection{Fast Preview with Noise Level Predictor}
\label{sec:edge}
On the edge, the lightweight GDM first generates a draft image \cite{c13} from the user's prompt to provide fast feedback. When the user refines the prompt, the edge model updates the image using an image-to-image (img2img) pipeline \cite{c12,c18}. As shown in \autoref{fig:img2img_edge}, to ensure efficiency, we introduce a semantic-aware noise level predictor that adapts the strength parameter based on the semantic difference between the previous and current prompts \cite{c16,c17}. This parameter controls the noise level, thus the number of diffusion steps, to perform on the latent image.

Specifically, we use a text encoder to extract semantic embeddings from prompts. Given previous and current prompts $p_{t-1}$ and $p_t$, their embeddings are computed as $\mathbf{h}_{t-1}=f_{\text{MiniLM}}(p_{t-1})$ and $\mathbf{h}_t=f_{\text{MiniLM}}(p_t)$, where $f_{\text{MiniLM}}(\cdot)$ maps a prompt to a $d$-dimensional embedding $\mathbf{h}\in\mathbb{R}^d$. A lightweight feed-forward network (FFN) then predicts the strength $\hat{s}_t = g\!\left([\mathbf{h}_{t-1},\,\mathbf{h}_t,\,\mathbf{h}_t-\mathbf{h}_{t-1}]\right)$, where $g(\cdot)$ is the FFN. To train the FFN, we construct a dataset that contains pairs of prompts, such as $(p_{t-1}, p_t)$ and their corresponding ground truth $s^*$. To be specific, we empirically predefine a discrete set of candidate strengths ranging from 0.40 to 0.90 with a step size of 0.05. Next, for each prompt pair $(p_{t-1}, p_t)$, we perform the img2img pipeline 
$I(\mathbf{x}_{t-1}, p_t; s)$ with different $s \in \mathcal{S}$, 
and compute the CLIP score between the generated image and the current prompt $p_t$, respectively. The strength that achieves the highest score is taken as the ground-truth label, given by
\begin{equation}
s_t^{*} = \arg\max_{s \in \mathcal{S}} \ \text{CLIP}\bigg(I(\mathbf{x}_{t-1}, p_t; s),\, p_t\bigg),
\end{equation}
where $\text{CLIP}(\cdot, \cdot)$ computes the image–text alignment score, and $\mathbf{x}_{t-1}$ is the image generated in the previous round. Thus, the loss function for training the strength predictor can be expressed as
\begin{equation}
\mathcal{L}_{\text{edge}} = \frac{1}{N} \sum_{n=1}^{N} \left( \hat{s}_{t} - s^{*}_{t} \right)^{2}
\end{equation}
where $N$ is the number of training pairs. We note that during training and inference, the predicted strength $\hat{s}_t$ is clipped to satisfy the range of the predefined candidate set $\mathcal{S}$.

\begin{figure}[t]
  \centering
  \includegraphics[width=\columnwidth]{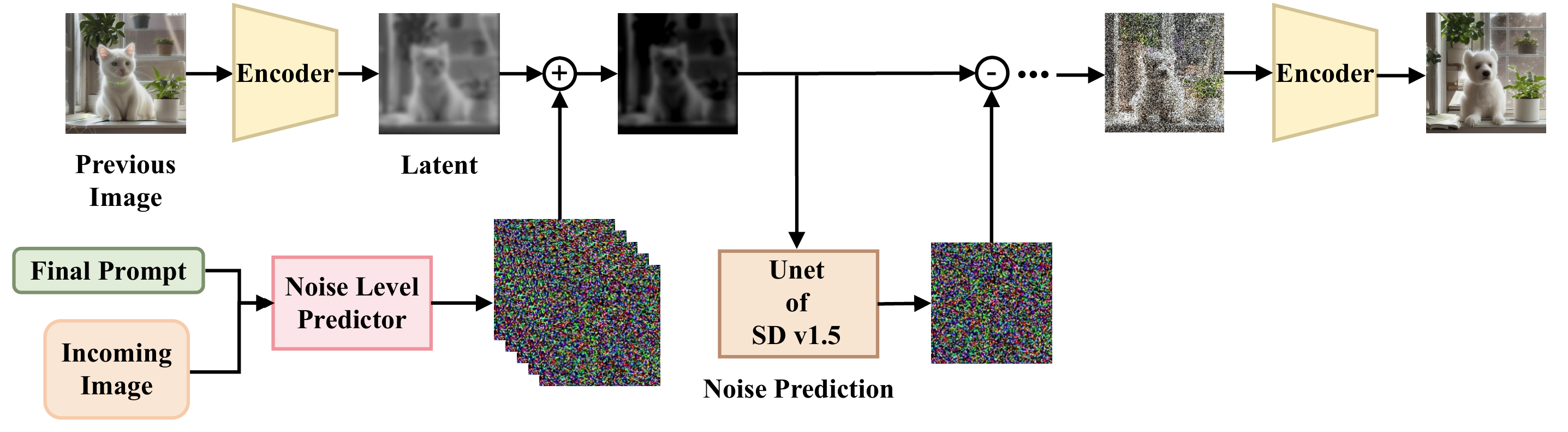}
  \caption{The structure of img2img generation on the cloud with a noise level predictor for refinement.}
  \label{fig:img2img_cloud}
\end{figure}

\subsection{High-Fidelity Noise Level Predictor with Skip-Step Acceleration}
\label{sec:cloud_strength}
Once the user finalizes the prompt, the edge sends the draft image and confirmed prompt to the cloud for high-quality refinement. Similar to the edge, the cloud uses an img2img pipeline and a strength predictor to avoid redundant denoising, but with higher capacity and multimodal fusion to improve quality, as shown in \autoref{fig:img2img_cloud}. The cloud employs a Unet architecture, derived from SD v1.5, to extract noise, predict the noise level, and perform denoising to refine the generated image.

To train this predictor, we use a higher-capacity language encoder, such as BERT \cite{c26}, to extract semantic embeddings $\mathbf{h}_{\text{cloud}}$ from the prompt, and the CLIP image encoder to obtain visual embeddings $\mathbf{v}_\text{cloud}$ from the draft image. These embeddings are concatenated to form a multimodal feature $\mathbf{z}_{\text{cloud}} = \phi(\mathbf{h}_{\text{cloud}}, \mathbf{v}_\text{cloud})$, where $\phi(\cdot)$ denotes concatenation. A deep regression head then maps this feature to a continuous strength value $\hat{s}_t^{\text{cloud}}$ controlling the noise level in the img2img pipeline:
\begin{equation}
\hat{s}_{\text{cloud}} = f_{\text{DeepReg}}\!\left(\mathbf{h}_{\text{cloud}}, \mathbf{v}_\text{cloud}\right),
\label{eq:cloud_pred}
\end{equation}
where $f_{\text{DeepReg}}(\cdot)$ is the regression head. 

The cloud predictor is trained similarly to the edge predictor, using a prompt dataset and corresponding CLIP-selected strengths as ground truth. The loss function is:
\begin{equation}
\mathcal{L}_{\text{cloud}} = \frac{1}{N} \sum_{t=1}^{N} \left( \hat{s}_{t}^{\text{cloud}} - s_{t}^{*} \right)^{2} + \lambda \, \Omega(\theta),
\label{eq:cloud_loss}
\end{equation}
where $s^{*}_{\text{cloud}}$ is the CLIP-derived strength, $\hat{s}^{\text{cloud}}$ is the predicted strength, $\Omega(\theta)$ is the regularization term, and $\lambda>0$ balances regularization and regression loss. Based on the predicted strength, the cloud performs img2img refinement with a skip-step denoising schedule to reduce redundant computation.




\vspace{-1em}
\begin{table}[t!]
\centering
\resizebox{\columnwidth}{!}{
\begin{tabular}{cccc}
\toprule
\textbf{Model} & \textbf{FID ↓} & \textbf{CLIP SCORE ↑} & \textbf{IS (SD) ↑} \\
\midrule
SD v1.5 & \textbf{12.808} & \underline{0.297} & \underline{23.387} (1.809) \\
Tiny-SD & 45.482 & 0.249 & 13.800 (0.599) \\
DiffusionX & \underline{17.016} & \textbf{0.313} & \textbf{24.943} (2.166) \\
\bottomrule
\end{tabular}
}
\caption{Comparison of image generation quality across different models in terms of FID, CLIP, and IS SCORE on the MS-COCO 30K Dataset.\vspace{-0.3em}}
\label{tab:image_quality_comparison}
\end{table}
\section{EXPERIMENTS}
\label{sec:EXP}

\subsection{Experimental Setup}

In experiments, we evaluate the peformance of the proposed \textit{DiffusionX} and also provide two baselines for comparison: 1) \textit{Tiny SD} deployed on the edge and 2) \textit{Stable Diffusion} v1.5 (SD v1.5) on the cloud. The edge equipped with an NVIDIA RTX 4060 8GB GPU, while the cloud uses an NVIDIA RTX A6000 48GB GPU. We assess image quality using the MS-COCO 30K dataset \cite{linMicrosoftCOCOCommon2014}, reporting FID \cite{c27}, CLIP Score, and IS \cite{c28}. To evaluate generation speed, we construct the COCO2017-Interactive-Prompts-400 dataset, based on the COCO 2017 dataset \cite{linMicrosoftCOCOCommon2014}, which simulates user interactions by progressively updating prompt pairs with 400 captions. As for the connection between the edge and cloud, we assume an uplink bandwidth of 20 Mbps, which is typical for 4G/5G networks \cite{bandwidth}. 
\begin{figure*}[t]
  \centering
  \includegraphics[width=\textwidth]{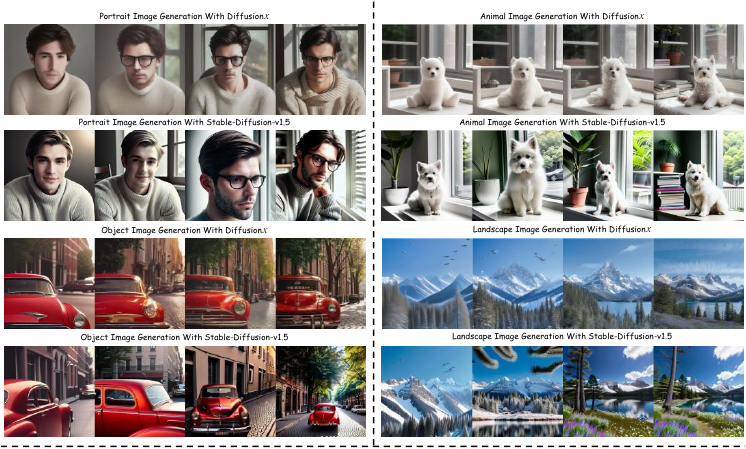}
  \caption{Visual examples of images generated by SD v1.5 and the proposed \textit{DiffusionX} on the COCO2017-Interactive-Prompts-400 Dataset.\vspace{-1em}}
  \label{fig:pinjietu}
\end{figure*}

\vspace{-0.2em}
\subsection{Image Generation Quality}

As shown in \autoref{tab:image_quality_comparison}, we compare the image generation quality of the three models on the MS-COCO 30K dataset in terms of FID, CLIP score, and IS. We can see that the proposed \textit{DiffusionX} achieves the highest average CLIP score of 0.313, indicating the best alignment between generated images and textual prompts among the compared models. Moreover, the proposed \textit{DiffusionX} also achieves comparable FID and IS scores to SD v1.5, and outperforms Tiny-SD significantly in all metrics. These results demonstrate the effectiveness of the proposed \textit{DiffusionX}.

As shown in \autoref{fig:pinjietu}, we provide some visual examples of images generated by SD v1.5 and the proposed \textit{DiffusionX}. We can see that both models can generate high-quality images that align well with the prompts. These visual results further validate the effectiveness of the proposed \textit{DiffusionX} in generating high-quality images from textual prompts.

\vspace{-1em}

\subsection{Image Generation Efficiency}

As shown in \autoref{tab:time_comparison}, we compare the image generation latency of the three models on the COCO2017-Interactive-Prompts-400 dataset. We can see that the proposed \textit{DiffusionX} can reduce the average total generation time by 2.23s compared to SD v1.5 running on the cloud, while being only 0.13s slower than Tiny-SD running on the edge. We note that although \textit{DiffusionX} introduces an additional transmission, the incurred latency is only 0.20s, which accounts for a small fraction of the total system latency.. These demonstrates that the proposed \textit{DiffusionX} can achieve a good balance between efficiency and image quality. 
\begin{table}[t]
\centering
\resizebox{\columnwidth}{!}{
\begin{tabular}{cccc}
\toprule
\textbf{Model} & \textbf{Trans. Latency (s) ↓ } & \textbf{Total Latency (s) ↓} \\
\midrule
SD v1.5 & -   & 14.15 \\
Tiny-SD               & -   & \textbf{11.79} \\
DiffusionX      & 0.20 & \underline{11.92} \\
\bottomrule
\end{tabular}
}
\caption{Efficiency comparison across different models in terms of average transmission latency and total latency on the COCO2017-Interactive-Prompts-400 Dataset.\vspace{-1em}}
\label{tab:time_comparison}
\end{table}
\begin{table}[H]
\centering
\resizebox{\columnwidth}{!}{
\begin{tabular}{cccc}
\toprule
\textbf{Model}  & \textbf{Trans. Latency (s) ↓} & \textbf{Avg. Total Time (s) ↓} \\
\midrule
w/o predictor & 0.20 & 13.96 \\
w/ predictor       & 0.20 & \textbf{11.92} (-15.8\%) \\
\bottomrule
\end{tabular}
}
\caption{Ablation study on the impact of the noise level predictor on latency.\vspace{-1em}}
\label{tab:time_comparison_xiao}
\end{table}

\subsection{Ablation Studies}
We also conduct an ablation study to assess the impact of the proposed noise level predictor on the performance of \textit{DiffusionX}. As shown in \autoref{tab:time_comparison_xiao}, we frist compare the system latency of \textit{DiffusionX} with and without the predictor. We can see that adding the predictor reduces the average total generation time from 13.96s to 11.92s, achieving a 15.8\% speedup. This is because the predictor helps avoid redundant denoising steps, thereby reducing computation.
\begin{table}[t]
\centering
\resizebox{\columnwidth}{!}{
\begin{tabular}{ccccc}
\toprule
\textbf{Model} & \textbf{FID ↓} &   \textbf{CLIP SCORE ↑}  & \textbf{IS (SD) ↑} \\
\midrule
w/o predictor& 21.453 &  \textbf{0.318} & \textbf{25.399} (2.851) \\
w/ predictor & \textbf{17.016} &  0.313  & 24.943 (2.166) \\
\bottomrule
\end{tabular}
}
\caption{Ablation study on the impact of the noise level predictor on image generation quality.\vspace{-1em}}
\label{tab:image_quality_comparison_xiao}
\end{table}

Moreover, we compare the image generation quality of \textit{DiffusionX} with and without the predictor on the MS-COCO 30K dataset, as shown in \autoref{tab:image_quality_comparison_xiao}. We can see that adding the predictor reduces FID from 21.453 to 17.016, while maintaining competitive CLIP and IS scores. This indicates that the proposed predictor can help improve image quality while reducing system latency as well. These results demonstrate the effectiveness of the proposed noise level predictor.

\section{CONCLUSIONS}
\label{sec:conclusions}
In this paper, we proposed \textit{DiffusionX}, a cloud–edge collaborative framework for efficient multi-round text-to-image generation, where the edge provides fast previews for user interaction and the cloud refines them into high-fidelity results. We introduced strength predictors on both sides to reduce redundant noise estimation, thereby lowering system overhead and accelerating iterative refinement. Extensive experiments demonstrated that the proposed \textit{DiffusionX} can reduce average total generation latency by over 15.8\% compared to SD v1.5 , while maintaining competitive image quality. 

\section{ACKNOWLEDGEMENT}
\label{sec:acknowledge}
This work is partly supported by the NSFC under grant No. 62293481, No. 62571487, No. 62201505, by the National Key R\&D Program of China under Grant 2024YFE0200802, and by the Zhejiang Provincial Natural Science Foundation of China under Grant No. LZ25F010001.

\vfill\pagebreak


{\renewcommand{\baselinestretch}{0.9}\selectfont
\small
\bibliographystyle{IEEEbib}
\bibliography{refs}
}

\end{document}